\pdfoutput=1

\documentclass[11pt]{article}

\usepackage{acl}

\usepackage{times}
\usepackage{latexsym}

\usepackage[T1]{fontenc}

\usepackage[utf8]{inputenc}

\usepackage{microtype}

\usepackage{graphicx}
\usepackage{CJKutf8}
\usepackage{stfloats}
\usepackage{multirow}
\usepackage{bm}
\usepackage{amsmath}
\usepackage{array}
\usepackage{booktabs}
\usepackage{makecell}
\usepackage{pifont}

\newcommand{\PreserveBackslash}[1]{\let\temp=\\#1\let\\=\temp}
\newcolumntype{C}[1]{>{\PreserveBackslash\centering}p{#1}}
\newcolumntype{R}[1]{>{\PreserveBackslash\raggedleft}p{#1}}
\newcolumntype{L}[1]{>{\PreserveBackslash\raggedright}p{#1}}

\newcommand{\cmark}{\ding{51}}
\newcommand{\xmark}{\ding{55}}
\DeclareMathOperator{\crossattn}{Cross-Attn}
\DeclareMathOperator{\softmax}{softmax}

\title{BLCU-ICALL at SemEval-2022 Task 1: Cross-Attention Multitasking Framework for Definition Modeling}

\author{Cunliang Kong, Yujie Wang, Ruining Chong, Liner Yang, Hengyuan Zhang, Erhong Yang, Yaping Huang}

\author{Cunliang Kong\textmd{\textsuperscript{1}},
	Yujie Wang\textmd{\textsuperscript{2}},
	Ruining Chong\textmd{\textsuperscript{1}},
	Liner Yang\textmd{\textsuperscript{1}\Thanks{~Corresponding author: Liner Yang.}}, \\
	\textbf{Hengyuan Zhang\textmd{\textsuperscript{1}},
		Erhong Yang\textmd{\textsuperscript{1}},
		Yaping Huang\textmd{\textsuperscript{2}}
	} \\
	\textsuperscript{1}School of Information Science, Beijing Language and Culture University \\
	\textsuperscript{2}School of Computer and Information Technology, Beijing Jiaotong University \\
    \texttt{cunliang.kong@outlook.com}
}

\begin{document}
\maketitle
\begin{abstract}
This paper describes the BLCU-ICALL system used in the SemEval-2022 Task 1 Comparing Dictionaries and Word Embeddings, the Definition Modeling subtrack, achieving 1st on Italian, 2nd on Spanish and Russian, and 3rd on English and French.
We propose a transformer-based multitasking framework to explore the task.
The framework integrates multiple embedding architectures through the cross-attention mechanism, and captures the structure of glosses through a masking language model objective.
Additionally, we also investigate a simple but effective model ensembling strategy to further improve the robustness.
The evaluation results show the effectiveness of our solution.
We release our code at: \href{https://github.com/blcuicall/SemEval2022-Task1-DM}{https://github.com/ blcuicall/SemEval2022-Task1-DM}.
\end{abstract}

\section{Introduction}
Word embeddings \cite{mikolov-2013-efficient,pennington-2014-glove,yogatama-2015-learning} are dense and low dimensional vectors used in many NLP tasks because they are found to be useful representations of words and often lead to better performance in various tasks.
In recent years, large pretrained language models (PLMs), such as BERT \cite{devlin-2019-bert} and GPT \cite{petroni-2019-language} families of models, have taken the NLP field by storm, achieving state-of-the-art performance on many tasks \cite{min-2021-recent}.
The contextual embeddings generated by PLMs are proven to capture syntax and semantic features of words \cite{jawahar-2019-does, turton-2020-deriving}.
But for human beings, word embeddings containing these information is still a \textit{black box} and unexplainable. 

There have been many efforts devoted to evaluating the word embeddings' lexical information, such as the word similarity \cite{landauer-1997-solution, downey-2007-sparse} and analogical relation \cite{mikolov-2013-linguistic} tasks.
However, these tasks can only serve as indirect evaluation methods. 
In light of this, \citet{noraset-2017-definition} proposed the task of definition modeling to evaluate whether a word embedding can be employed to generate a dictionary gloss.
Since the gloss is a direct and explicit statement of word meaning, this task provides a more transparent view.

\begin{figure}[t]
	\centering
	\includegraphics[width=\linewidth]{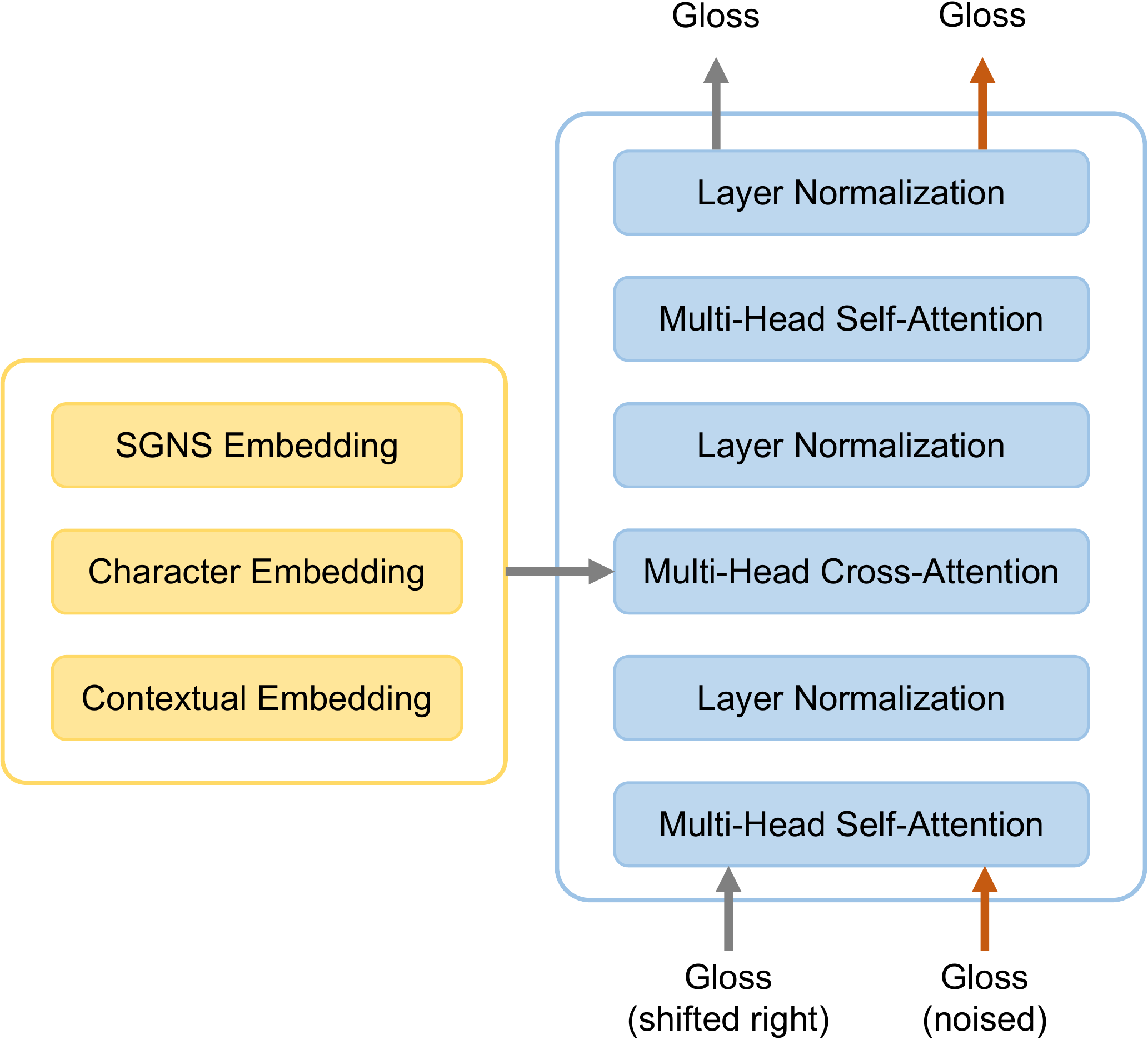}
	\caption{
		Architecture of the Cross-Attention Multitasking Framework.
	}
	\label{fig:arch}
\end{figure}

The SemEval-2022 Task 1 Comparing Dictionaries and Word Embeddings \cite{mickus-etal-2022-semeval} aims at comparing the two types of semantic descriptions: dictionary glosses and word embeddings.
The subtrack 1 is a definition modeling task, which requires models to generate glosses from word embeddings.
The task provides data from 5 languages (English, Spanish, French, Italian, Russian) as well as static, character, and contextual embeddings.

Our team propose a transformer-based \cite{vaswani-2017-attention} Cross-Attention Multitasking Framework to explore the task and apply the framework to all 5 languages.
We integrate the multiple embedding architectures through a cross-attention mechanism, which allows the model to query all the embeddings at each time step during generation.
To better capture the structure of glosses, we employ an additional masking language model (MLM) \cite{devlin-2019-bert} into the framework.
We also investigate the ensemble strategies to further enhance the robustness.

Therefore, the contributions of our system lie in:
\begin{itemize}
	\item
	We propose the Cross-Attention Multitasking Framework as a novel solution to the definition modeling task.
	\item
	The evaluation results show the effectiveness of our solution.
	Our system achieves 1st on Italian, 2nd on Spanish and Russian, and 3rd on English and French.
\end{itemize}

\begin{table*}[t]
	\centering
	\begin{tabular}{lR{1.5cm}R{1.5cm}R{1.5cm}R{1.5cm}R{1.5cm}R{1.5cm}R{1.5cm}}
		\toprule
		& Train & Dev. & Test & SGNS Emb. & Character Emb. & Electra Emb. & Gloss Len. \\
		\midrule
		English & 43,608 & 6,375 & 6,221 & \cmark & \cmark & \cmark & 11.73 \\
		Spanish & 43,608 & 6,375 & 6,221 & \cmark & \cmark & \xmark & 14.84 \\
		French & 43,608 & 6,375 & 6,221 & \cmark & \cmark & \cmark & 14.31 \\
		Italian & 43,608 & 6,375 & 6,221 & \cmark & \cmark & \xmark & 13.58 \\
		Russian & 43,608 & 6,375 & 6,221 & \cmark & \cmark & \cmark & 11.32 \\
		\bottomrule	
	\end{tabular}
	\caption{Detailed statistics of the dataset. The last column lists the average length of glosses in the training set.}
	\label{table:dataset}
\end{table*}

\section{Background}
The definition modeling subtrack provides participants with a multilingual dataset in the form of $\{E, \bm g\}$, where $E$ is a set including SGNS \cite{mikolov-2013-distributed}, character \cite{kim-2016-character}, and Electra \cite{clark-2020-electra} embeddings, and $\bm g$ is a dictionary gloss.
This task takes $E$ as the input, and requires models to generate $\bm g$.
Note that all the embeddings have 256 dimensions, and the Electra embeddings are only available for 3 of the 5 languages.
More detailed statistics of the dataset are listed in Table \ref{table:dataset}.

Many previous work used additional data to improve the performance of generation, such as example sentences \cite{gadetsky-2018-conditional, chang-2018-xsense, ishiwatari-2019-learning, kong-2020-toward} and semantic features \cite{yang-2020-incorp}.
Some studies also investigated how to employ PLMs for this task \cite{reid-2020-vcdm, bevilacqua-2020-generationary, huang-2021-definition, kong-etal-2022-simpdefiner}.

Differently, to keep the results linguistically significant and easily comparable, the SemEval-2022 Task 1 prohibits the usage of external data and PLMs.
Therefore, our system focuses on effectively integrating all given embeddings and  modeling the glosses.

\section{System Overview}
Figure \ref{fig:arch} illustrates the entire architecture of our system, which is a Cross-Attention Multitasking Framework based on transformer.
The framework consists of two objectives, namely the generation and reconstruction objectives.
This section introduces the system in detail.

\subsection{The Generation Objective}
The generation objective serves as a standard transformer decoder, which generates the gloss as the following language model:
\begin{equation}
	P(\bm g| E; \bm \theta) = \prod_{t} P\left(\bm g_{t} | \bm g_{<t}, E; \bm \theta \right),
\end{equation}
where $\bm g_t$ is the $t$-th token in the gloss, and $\bm \theta$ is the set of parameters.
The model is then optimized using the following loss function:
\begin{equation}
	\mathcal{L}_{gen}(\bm \theta) = -\sum_{\bm g \in D} \log P(\bm g | E; \bm \theta),
\end{equation}
where $D$ is the training dataset.

In the above operations, a crucial challenge is to integrate multiple embeddings corresponding to one word.
We assume that the SGNS, character, and Electra embeddings contain different lexical features, and better results can be obtained by comprehensively considering all the information.
To achieve that, we feed the set $E$, including all these embeddings, into the cross-attention mechanism:
\begin{equation}
	\crossattn(H, E, E) = \softmax(\frac{HE^{T}}{\sqrt{d_h}})E
\end{equation}
where $H$ is the hidden-states obtained from by self-attention, and $d_h$ is the dimension of the hidden-states.
This operation ensures the given embeddings are adaptively integrated at each time-step.

\subsection{The Reconstruction Objective}
Our system is a language model specially designed for dictionary glosses.
We further enhance this model by incorporating a reconstruction objective.

We corrupt each gloss $\bm g$  by randomly substituting or blanking some words.
And then we obtain a corrupted version $\tilde{\bm g}$.
We input $\tilde{\bm g}$ into our system and obtain $\bm g$ by solving a self-supervised task of:
\begin{equation}
	P(\bm g | \tilde{\bm g}; \bm \theta) = \prod_{t} P(\bm g_t | \bm g_{<t}, \tilde{\bm g}; \bm \theta).
\end{equation}
Note that we share exactly the same parameters $\bm \theta$ as in the generation objective.
The model is optimized by the following loss function:
\begin{equation}
		\mathcal{L}_{rec}(\bm \theta) = -\sum_{\bm g \in D} \log P(\bm g | \tilde{\bm g}; \bm \theta),
\end{equation}

The goal of the reconstruction objective is to better model the glosses.
Therefore, we don't use the given embeddings in this operation.
In practice, we feed a zero vector into the cross-attention mechanism to mask it out as $\crossattn(H, \bm 0, \bm 0)$.

\subsection{Training and Ensembling}
We train the entire multitasking framework by jointly minimizing the weighted sum of both loss functions:
\begin{equation}
	\mathcal{L} = \mathcal{L}_{gen} + \lambda\mathcal{L}_{rec},
	\label{eq:loss}
\end{equation}
where $\lambda$ is a hyper-parameter.

Model ensembling is proven to be effective to improve the robustness \cite{allen-2020-towards}.
In our work, we adopt a simple but effective model ensembling strategy.
We train a series of models initialized by different random seeds, and then vote with the trained models during inference.

\section{Experimental Setup}
\subsection{Implementation Details}
Many neural network-based generation systems struggle with the OOV (out-of-vocabulary) problem.
To alleviate the problem, we apply the SentencePiece algorithm \cite{kudo-richardson-2018-sentencepiece} to glosses to reduce the vocabulary size.
We use the tokenizers\footnote{tokenizers: \href{https://github.com/huggingface/tokenizers}{https://github.com/huggingface/tokenizers}.} toolkit for implementation and set the size to 10k for all 5 languages.  

Our system is a 3-layer, 8-head transformer-based model implemented by the Pytorch library \cite{paszke-2019-pytorch}.
We use the Adam optimizer \cite{kingma-2015-adam} with $\beta_1 = 0.9$, $\beta_2=0.98$ and $\epsilon=10^{-9}$.
We adopt the Noam Optimizer proposed by \citep{vaswani-2017-attention} with an initial learning rate of $1e\!-\!7$, a maximum learning rate of $1e\!-\!3$, and a minimum learning rate of $1e\!-\!9$.
We set the warmup steps to 4000 and batch size to 128.
The maximum epochs is set to 500.
And we set an early stop strategy in the patience of 5 epochs.
To avoid gradient exploding, we clipped the gradient norm within 0.1.
We also employ label smoothing technique \cite{pereyra-2017-regulariz} with a smoothing value of 0.1 during training.
For the gloss corruption in the reconstruction objective, we follow \citet{devlin-2019-bert} to randomly delete and blank words with a uniform probability of 0.2.
And the $\lambda$ (in Equation \ref{eq:loss}) is set to 1.
For model ensembling, we train 5 models with different seeds.
Due to the time constraints, our official submission has a result of ensembling three models on English, and results of single models on the reset of 4 languages.
We submitted the results of ensembling 5 models in the post-evaluation phase.


For each language, we use the development set released by organizers for model selection.
We select the best epoch using the summary of BLEU \cite{papineni-2002-bleu} and MoverScore \cite{zhao-2019-mover} on the development set.

\subsection{Evaluation Metrics}
The definition modeling subtrack uses three metrics, which are MoverScore \cite{zhao-2019-mover}, BLEU \cite{papineni-2002-bleu} , and lemma-level BLEU respectively.
Readers can refer to the task paper \cite{mickus-etal-2022-semeval} for more details.

\begin{table}[!t]
	\centering
	\small
	\begin{tabular}{l|lrrr}
		\toprule
		& Models & S-BLEU & L-BLEU & MvSc.  \\
		\midrule
		\multirow{5}{*}{EN} & SGNS & 0.00125 & 0.00250 & 0.10339 \\
		& Char & 0.00011 & 0.00022 & 0.08852 \\
		& Electra & 0.00165 & 0.00215 & 0.08798 \\
		& CAMF & \textbf{0.03127} & \textbf{0.03957} & \textbf{0.13475} \\
		& Ensemble & \underline{0.03106} & \underline{0.03906} & \underline{0.13273} \\
		\midrule
		\multirow{4}{*}{ES} & SGNS & 0.01536 & 0.02667 & \textbf{0.20130} \\
		& Char & 0.01505 & 0.02471 & \underline{0.19933} \\
		& CAMF & \underline{0.03914} & \underline{0.05606} & 0.12778 \\
		& Ensemble & \textbf{0.03925} & \textbf{0.05624} & 0.13121 \\
		\midrule
		\multirow{5}{*}{FR} & SGNS & 0.00351 & 0.00604 & \underline{0.18478} \\
		& Char & 0.00280 & 0.00706 & \textbf{0.18579} \\
		& Electra & 0.00219 & 0.00301 & 0.17391 \\
		& CAMF & \underline{0.02679} & \underline{0.03691} & 0.04193 \\
		& Ensemble & \textbf{0.02700} & \textbf{0.03738} & 0.04455\\
		\midrule
		\multirow{4}{*}{IT} & SGNS & 0.02591 & 0.04081 & \textbf{0.20527} \\
		& Char & 0.00640 & 0.00919 & \underline{0.15920} \\
		& CAMF & \underline{0.06646} & \underline{0.09926} & 0.11717 \\
		& Ensemble & \textbf{0.06812} & \textbf{0.10147} & 0.12233 \\
		\midrule
		\multirow{5}{*}{RU} & SGNS & 0.01520 & 0.02112 & \textbf{0.34716} \\
		& Char & 0.01313 & 0.01847 & 0.32307 \\
		& Electra & 0.01189 & 0.01457 & \underline{0.33577} \\
		& CAMF & \underline{0.04843} & \underline{0.06548} & 0.14820 \\
		& Ensemble & \textbf{0.05192} & \textbf{0.07074} & 0.15702 \\
		\bottomrule
	\end{tabular}
	\caption{Evaluation results of different models in 5 languages. The SGNS, Char, Electra are baseline models provided by the organizers. The CAMF (Cross-Attention Multilingual Framework) is the model of official submission. And the Ensemble is an ensemble of 5 models submitted in the post-evaluation. Bold and underline mark the best and second scores, respectively.}
	\label{table:main-results}
\end{table}

\section{Results and Analysis}
In this section, we present the evaluation results and discuss our analysis of the generated definitions.

\subsection{Main Results}
Table \ref{table:main-results} presents the evaluation scores on all 5 languages.
Results show that our system significantly outperforms the baseline models in terms of the sentence BLEU and lemma-level BLEU.
This indicates the effectiveness of our proposed cross-attention multitasking framework.
However, the SGNS and Char are strong baselines in terms of the MoverScore, and our system only outperforms the baselines on English.
We speculate that our results have more coincide words with references, but are not fluent enough, which leads to a low score from the pretrained model used by MoverScore.

We also observe that model ensembling has brought the improvement of performance.
It can be seen from the table that the Ensemble model outperforms the CAMF on 4 of the 5 languages, except for a slight decline on English.
This may be due to the randomness of the parameter initialization.
We also argue that better performance can be obtained by applying hyper-parameter searching algorithms and ensembling more models.

\begin{table}[t]
	\centering
	\small
	\begin{tabular}{L{1.2cm}|L{5.5cm}}
		\toprule
		\multicolumn{2}{l}{(1) Redundancy and overusing common phrases: 42.00\%} \\
		word & explosion \\
		reference & A sudden outburst. \\
		hypothesis &  A sudden, sudden, or destruction. \\
		\midrule
		\multicolumn{2}{l}{(2) Self-reference: 2.00\%}\\
		word & discover \\
		reference & To reveal (information); to divulge, make known. \\
		hypothesis &  To make a conclusion of; to discover. \\
		\midrule
		\multicolumn{2}{l}{(3) Wrong Part-Of-Speech: 5.50\%} \\
		word & genius \\
		reference & ingenious, brilliant, very clever, or original. \\
		hypothesis &  A person or thing that is extraordinary. \\
		\midrule
		\multicolumn{2}{l}{(4) Under-specified: 23.50\%} \\
		word & mayor \\
		reference & The leader of a city. \\
		hypothesis &  A person who is a member of authority. \\
		\midrule
		\multicolumn{2}{l}{(5) Opposite: 2.00\%} \\
		word & solid \\
		reference & Excellent , of high quality , or reliable. \\
		hypothesis &  Having no size or value. \\
		\midrule
		\multicolumn{2}{l}{(6) Close Semantics: 17.00\%} \\
		word & bed \\
		reference & The time for going to sleep or resting in bed. \\
		hypothesis &  The state or quality of being a room. \\
		\midrule 
		\multicolumn{2}{l}{(7) Incorrect: 52.00\%}\\
		word & smooth \\
		reference &  Lacking projections or indentations; not serrated. \\
		hypothesis &  Having the shape of a tree. \\
		\bottomrule
	\end{tabular}
	\caption{Error types and examples.}
	\label{table:error}
\end{table}

\subsection{Error Analysis}
In order to qualitatively analyze the definitions generated by our system, we randomly select several  items from the English test set and manually annotate the error types following \citet{noraset-2017-definition}.
In total, we extract 200 items, of which 197 contain some degree of error.
We illustrate the error types and examples in Table \ref{table:error}.
Note that each item may contain multiple errors, so the sum of the percentages in the table is greater than 100\%.

From the table, we observe that the quality of English definitions generated by our system still need to be improved.
Error types (1) to (3) are problems from the system, and types (4) to (6) are shortcomings in the embeddings.
As we can see, the former accounts for a much larger proportion than the latter.
The 52\% incorrectness indicated by type (7) shows that many glosses generated by our system are irrelevant to the word.
And the dataset released in this task will support significant future work on the definition modeling task.

\section{Conclusion}
In this paper, we present the implementation of the BLCU-ICALL system submitted to the SemEval-2022 Task 1, Definition Modeling subtrack.
We propose a Cross-Attention Multitasking Framework that leverages multiple embedding architectures and jointly trains two objectives.
We also investigate a simple but effective ensembling strategy to enhance the robustness.
In future efforts, we plan to further improve our system to better handle the problems of redundancy and incorrect glosses.

\section*{Acknowledgements}
This work was supported by the funds of Beijing Advanced Innovation Center for Language Resources (No. TYZ19005), Research Project of the National Language Commission (No. ZDI135-105, No. ZDI135-131), and BLCU supported project for young researchers program (supported by the Fundamental Research Funds for the Central Universities)(No. 20YCX142). We would like to thank all anonymous reviewers for their valuable comments and suggestions on this work.

\bibliography{custom}

\begin{thebibliography}{33}
\expandafter\ifx\csname natexlab\endcsname\relax\def\natexlab#1{#1}\fi

\bibitem[{Allen-Zhu and Li(2020)}]{allen-2020-towards}
Zeyuan Allen-Zhu and Yuanzhi Li. 2020.
\newblock Towards understanding ensemble, knowledge distillation and
  self-distillation in deep learning.
\newblock \emph{arXiv preprint arXiv:2012.09816}.

\bibitem[{Bevilacqua et~al.(2020)Bevilacqua, Maru, and
  Navigli}]{bevilacqua-2020-generationary}
Michele Bevilacqua, Marco Maru, and Roberto Navigli. 2020.
\newblock Generationary or {``}how we went beyond word sense inventories and
  learned to gloss{''}.
\newblock In \emph{Proceedings of the 2020 Conference on Empirical Methods in
  Natural Language Processing (EMNLP)}, pages 7207--7221, Online. Association
  for Computational Linguistics.

\bibitem[{Chang et~al.(2018)Chang, Chi, Tsai, and Chen}]{chang-2018-xsense}
Ting{-}Yun Chang, Ta{-}Chung Chi, Shang{-}Chi Tsai, and Yun{-}Nung Chen. 2018.
\newblock xsense: Learning sense-separated sparse representations and textual
  definitions for explainable word sense networks.
\newblock \emph{CoRR}, abs/1809.03348.

\bibitem[{Clark et~al.(2020)Clark, Luong, Le, and Manning}]{clark-2020-electra}
Kevin Clark, Minh-Thang Luong, Quoc~V Le, and Christopher~D Manning. 2020.
\newblock Electra: Pre-training text encoders as discriminators rather than
  generators.
\newblock \emph{arXiv preprint arXiv:2003.10555}.

\bibitem[{Devlin et~al.(2019)Devlin, Chang, Lee, and
  Toutanova}]{devlin-2019-bert}
Jacob Devlin, Ming-Wei Chang, Kenton Lee, and Kristina Toutanova. 2019.
\newblock {BERT}: Pre-training of deep bidirectional transformers for language
  understanding.
\newblock In \emph{Proceedings of the 2019 Conference of the North {A}merican
  Chapter of the Association for Computational Linguistics: Human Language
  Technologies, Volume 1 (Long and Short Papers)}, pages 4171--4186,
  Minneapolis, Minnesota. Association for Computational Linguistics.

\bibitem[{Downey et~al.(2007)Downey, Schoenmackers, and
  Etzioni}]{downey-2007-sparse}
Doug Downey, Stefan Schoenmackers, and Oren Etzioni. 2007.
\newblock Sparse information extraction: Unsupervised language models to the
  rescue.
\newblock In \emph{ACL}.

\bibitem[{Gadetsky et~al.(2018)Gadetsky, Yakubovskiy, and
  Vetrov}]{gadetsky-2018-conditional}
Artyom Gadetsky, Ilya Yakubovskiy, and Dmitry Vetrov. 2018.
\newblock Conditional generators of words definitions.
\newblock In \emph{Proceedings of the 56th Annual Meeting of the Association
  for Computational Linguistics (Volume 2: Short Papers)}, pages 266--271.

\bibitem[{Huang et~al.(2021)Huang, Kajiwara, and Arase}]{huang-2021-definition}
Han Huang, Tomoyuki Kajiwara, and Yuki Arase. 2021.
\newblock Definition modelling for appropriate specificity.
\newblock In \emph{Proceedings of the 2021 Conference on Empirical Methods in
  Natural Language Processing}, pages 2499--2509. Association for Computational
  Linguistics.

\bibitem[{Ishiwatari et~al.(2019)Ishiwatari, Hayashi, Yoshinaga, Neubig, Sato,
  Toyoda, and Kitsuregawa}]{ishiwatari-2019-learning}
Shonosuke Ishiwatari, Hiroaki Hayashi, Naoki Yoshinaga, Graham Neubig, Shoetsu
  Sato, Masashi Toyoda, and Masaru Kitsuregawa. 2019.
\newblock Learning to describe unknown phrases with local and global contexts.
\newblock In \emph{Proceedings of the 2019 Conference of the North {A}merican
  Chapter of the Association for Computational Linguistics: Human Language
  Technologies, Volume 1 (Long and Short Papers)}, pages 3467--3476.
  Association for Computational Linguistics.

\bibitem[{Jawahar et~al.(2019)Jawahar, Sagot, and Seddah}]{jawahar-2019-does}
Ganesh Jawahar, Beno{\^\i}t Sagot, and Djam{\'e} Seddah. 2019.
\newblock What does bert learn about the structure of language?
\newblock In \emph{ACL 2019-57th Annual Meeting of the Association for
  Computational Linguistics}.

\bibitem[{Kim et~al.(2016)Kim, Jernite, Sontag, and Rush}]{kim-2016-character}
Yoon Kim, Yacine Jernite, David Sontag, and Alexander~M Rush. 2016.
\newblock Character-aware neural language models.
\newblock In \emph{Thirtieth AAAI conference on artificial intelligence}.

\bibitem[{Kingma and Ba(2015)}]{kingma-2015-adam}
Diederik~P Kingma and Jimmy Ba. 2015.
\newblock Adam: A method for stochastic optimization.
\newblock In \emph{ICLR (Poster)}.

\bibitem[{Kong et~al.(2022)Kong, Chen, Zhang, Yang, and
  Yang}]{kong-etal-2022-simpdefiner}
Cunliang Kong, Yun Chen, Hengyuan Zhang, Liner Yang, and Erhong Yang. 2022.
\newblock Multitasking framework for unsupervised simple definition generation.
\newblock In \emph{Proceedings of the 60th Annual Meeting of the Association
  for Computational Linguistics}.

\bibitem[{Kong et~al.(2020)Kong, Yang, Zhang, Fan, Liu, Chen, and
  Yang}]{kong-2020-toward}
Cunliang Kong, Liner Yang, Tianzuo Zhang, Qinan Fan, Zhenghao Liu, Yun Chen,
  and Erhong Yang. 2020.
\newblock Toward cross-lingual definition generation for language learners.
\newblock \emph{arXiv preprint arXiv:2010.05533}.

\bibitem[{Kudo and Richardson(2018)}]{kudo-richardson-2018-sentencepiece}
Taku Kudo and John Richardson. 2018.
\newblock {S}entence{P}iece: A simple and language independent subword
  tokenizer and detokenizer for neural text processing.
\newblock In \emph{Proceedings of the 2018 Conference on Empirical Methods in
  Natural Language Processing: System Demonstrations}, pages 66--71, Brussels,
  Belgium. Association for Computational Linguistics.

\bibitem[{Landauer and Dumais(1997)}]{landauer-1997-solution}
Thomas~K Landauer and Susan~T Dumais. 1997.
\newblock A solution to plato's problem: The latent semantic analysis theory of
  acquisition, induction, and representation of knowledge.
\newblock \emph{Psychological review}, 104(2):211.

\bibitem[{Mickus et~al.(2022)Mickus, Paperno, Constant, and van
  Deemter}]{mickus-etal-2022-semeval}
Timothee Mickus, Denis Paperno, Mathieu Constant, and Kees van Deemter. 2022.
\newblock {SemEval-2022 Task 1}: Codwoe -- comparing dictionaries and word
  embeddings.
\newblock In \emph{Proceedings of the 16th International Workshop on Semantic
  Evaluation (SemEval-2022)}. Association for Computational Linguistics.

\bibitem[{Mikolov et~al.(2013{\natexlab{a}})Mikolov, Chen, Corrado, and
  Dean}]{mikolov-2013-efficient}
Tomas Mikolov, Kai Chen, Greg Corrado, and Jeffrey Dean. 2013{\natexlab{a}}.
\newblock Efficient estimation of word representations in vector space.
\newblock \emph{arXiv preprint arXiv:1301.3781}.

\bibitem[{Mikolov et~al.(2013{\natexlab{b}})Mikolov, Sutskever, Chen, Corrado,
  and Dean}]{mikolov-2013-distributed}
Tomas Mikolov, Ilya Sutskever, Kai Chen, Greg~S Corrado, and Jeff Dean.
  2013{\natexlab{b}}.
\newblock Distributed representations of words and phrases and their
  compositionality.
\newblock \emph{Advances in neural information processing systems}, 26.

\bibitem[{Mikolov et~al.(2013{\natexlab{c}})Mikolov, Yih, and
  Zweig}]{mikolov-2013-linguistic}
Tom{\'a}{\v{s}} Mikolov, Wen-tau Yih, and Geoffrey Zweig. 2013{\natexlab{c}}.
\newblock Linguistic regularities in continuous space word representations.
\newblock In \emph{Proceedings of the 2013 conference of the north american
  chapter of the association for computational linguistics: Human language
  technologies}, pages 746--751.

\bibitem[{Min et~al.(2021)Min, Ross, Sulem, Veyseh, Nguyen, Sainz, Agirre,
  Heinz, and Roth}]{min-2021-recent}
Bonan Min, Hayley Ross, Elior Sulem, Amir Pouran~Ben Veyseh, Thien~Huu Nguyen,
  Oscar Sainz, Eneko Agirre, Ilana Heinz, and Dan Roth. 2021.
\newblock Recent advances in natural language processing via large pre-trained
  language models: A survey.
\newblock \emph{arXiv preprint arXiv:2111.01243}.

\bibitem[{Noraset et~al.(2017)Noraset, Liang, Birnbaum, and
  Downey}]{noraset-2017-definition}
Thanapon Noraset, Chen Liang, Larry Birnbaum, and Doug Downey. 2017.
\newblock Definition modeling: Learning to define word embeddings in natural
  language.
\newblock In \emph{Thirty-First AAAI Conference on Artificial Intelligence}.

\bibitem[{Papineni et~al.(2002)Papineni, Roukos, Ward, and
  Zhu}]{papineni-2002-bleu}
Kishore Papineni, Salim Roukos, Todd Ward, and Wei-Jing Zhu. 2002.
\newblock Bleu: a method for automatic evaluation of machine translation.
\newblock In \emph{Proceedings of the 40th annual meeting of the Association
  for Computational Linguistics}, pages 311--318.

\bibitem[{Paszke et~al.(2019)Paszke, Gross, Massa, Lerer, Bradbury, Chanan,
  Killeen, Lin, Gimelshein, Antiga et~al.}]{paszke-2019-pytorch}
Adam Paszke, Sam Gross, Francisco Massa, Adam Lerer, James Bradbury, Gregory
  Chanan, Trevor Killeen, Zeming Lin, Natalia Gimelshein, Luca Antiga, et~al.
  2019.
\newblock Pytorch: An imperative style, high-performance deep learning library.
\newblock \emph{Advances in neural information processing systems}, 32.

\bibitem[{Pennington et~al.(2014)Pennington, Socher, and
  Manning}]{pennington-2014-glove}
Jeffrey Pennington, Richard Socher, and Christopher~D Manning. 2014.
\newblock Glove: Global vectors for word representation.
\newblock In \emph{Proceedings of the 2014 conference on empirical methods in
  natural language processing (EMNLP)}, pages 1532--1543.

\bibitem[{Pereyra et~al.(2017)Pereyra, Tucker, Chorowski, Kaiser, and
  Hinton}]{pereyra-2017-regulariz}
Gabriel Pereyra, George Tucker, Jan Chorowski, Lukasz Kaiser, and Geoffrey~E.
  Hinton. 2017.
\newblock Regularizing neural networks by penalizing confident output
  distributions.
\newblock \emph{CoRR}, abs/1701.06548.

\bibitem[{Petroni et~al.(2019)Petroni, Rockt{\"a}schel, Riedel, Lewis, Bakhtin,
  Wu, and Miller}]{petroni-2019-language}
Fabio Petroni, Tim Rockt{\"a}schel, Sebastian Riedel, Patrick Lewis, Anton
  Bakhtin, Yuxiang Wu, and Alexander Miller. 2019.
\newblock Language models as knowledge bases?
\newblock In \emph{Proceedings of the 2019 Conference on Empirical Methods in
  Natural Language Processing and the 9th International Joint Conference on
  Natural Language Processing (EMNLP-IJCNLP)}, pages 2463--2473, Hong Kong,
  China. Association for Computational Linguistics.

\bibitem[{Reid et~al.(2020)Reid, Marrese-Taylor, and Matsuo}]{reid-2020-vcdm}
Machel Reid, Edison Marrese-Taylor, and Yutaka Matsuo. 2020.
\newblock {VCDM: Leveraging Variational Bi-encoding and Deep Contextualized
  Word Representations for Improved Definition Modeling}.
\newblock In \emph{Proceedings of the 2020 Conference on Empirical Methods in
  Natural Language Processing (EMNLP)}, pages 6331--6344.

\bibitem[{Turton et~al.(2020)Turton, Vinson, and Smith}]{turton-2020-deriving}
Jacob Turton, David Vinson, and Robert~Elliott Smith. 2020.
\newblock Deriving contextualised semantic features from bert (and other
  transformer model) embeddings.
\newblock \emph{arXiv preprint arXiv:2012.15353}.

\bibitem[{Vaswani et~al.(2017)Vaswani, Shazeer, Parmar, Uszkoreit, Jones,
  Gomez, Kaiser, and Polosukhin}]{vaswani-2017-attention}
Ashish Vaswani, Noam Shazeer, Niki Parmar, Jakob Uszkoreit, Llion Jones,
  Aidan~N Gomez, {\L}ukasz Kaiser, and Illia Polosukhin. 2017.
\newblock Attention is all you need.
\newblock \emph{Advances in neural information processing systems}, 30.

\bibitem[{Yang et~al.(2020)Yang, Kong, Chen, Liu, Fan, and
  Yang}]{yang-2020-incorp}
Liner Yang, Cunliang Kong, Yun Chen, Yang Liu, Qinan Fan, and Erhong Yang.
  2020.
\newblock Incorporating sememes into chinese definition modeling.
\newblock \emph{IEEE/ACM Transactions on Audio, Speech, and Language
  Processing}, 28:1669--1677.

\bibitem[{Yogatama et~al.(2015)Yogatama, Faruqui, Dyer, and
  Smith}]{yogatama-2015-learning}
Dani Yogatama, Manaal Faruqui, Chris Dyer, and Noah Smith. 2015.
\newblock Learning word representations with hierarchical sparse coding.
\newblock In \emph{International Conference on Machine Learning}, pages 87--96.
  PMLR.

\bibitem[{Zhao et~al.(2019)Zhao, Peyrard, Liu, Gao, Meyer, and
  Eger}]{zhao-2019-mover}
Wei Zhao, Maxime Peyrard, Fei Liu, Yang Gao, Christian~M. Meyer, and Steffen
  Eger. 2019.
\newblock Moverscore: Text generation evaluating with contextualized embeddings
  and earth mover distance.
\newblock \emph{CoRR}, abs/1909.02622.

\end{thebibliography}
\bibliographystyle{acl_natbib}
\end{document}